\documentclass[letterpaper, 10 pt, conference]{ieeeconf}  

\IEEEoverridecommandlockouts                              

\usepackage{graphics} 
\usepackage{epsfig} 
\usepackage{times} 
\usepackage{amsmath} 
\usepackage{amssymb}  
\usepackage{xcolor}

\usepackage{multicol}
\usepackage[bookmarks=true]{hyperref}

\usepackage{cite}
\usepackage{algorithm, algpseudocode}
\let\labelindent\relax
\usepackage{enumitem}

\DeclareMathOperator*{\argmin}{arg\,min}
\newtheorem{definition}{Definition}

\title{\LARGE \bf Volumetric Ergodic Control}

\author{Jueun Kwon, Max M. Sun, and Todd Murphey
\thanks{The authors are with the Department of Mechanical Engineering, Northwestern University, Evanston, IL 60208, USA. Email: {\tt\small jueun@u.northwestern.edu}. This work is supported by ARO grant W911NF-19-1-0233 and W911NF-22-1-0286. The views expressed are the authors’ and not necessarily those of the funders.}
}

\begin{document}
\allowdisplaybreaks

\maketitle
\thispagestyle{empty}
\pagestyle{empty}


\begin{abstract}

Ergodic control synthesizes optimal coverage behaviors over spatial distributions for nonlinear systems. However, existing formulations model the robot as a non-volumetric point, whereas in practice a robot interacts with the environment through its body and sensors with physical volume. In this work, we introduce a new ergodic control formulation that optimizes spatial coverage using a volumetric state representation. Our method preserves the asymptotic coverage guarantees of ergodic control, adds minimal computational overhead for real-time control, and supports arbitrary sample-based volumetric models. We evaluate our method across search and manipulation tasks---with multiple robot dynamics and end-effector geometries or sensor models---and show that it improves coverage efficiency by more than a factor of two while maintaining a $100\%$ task completion rate across all experiments, outperforming the standard ergodic control method. Finally, we demonstrate the effectiveness of our method on a robot arm performing mechanical erasing tasks. Project website: \url{https://murpheylab.github.io/vec/}
\end{abstract}


\begin{figure*}[h] 
    \centering    
    \includegraphics[width=\textwidth, keepaspectratio]{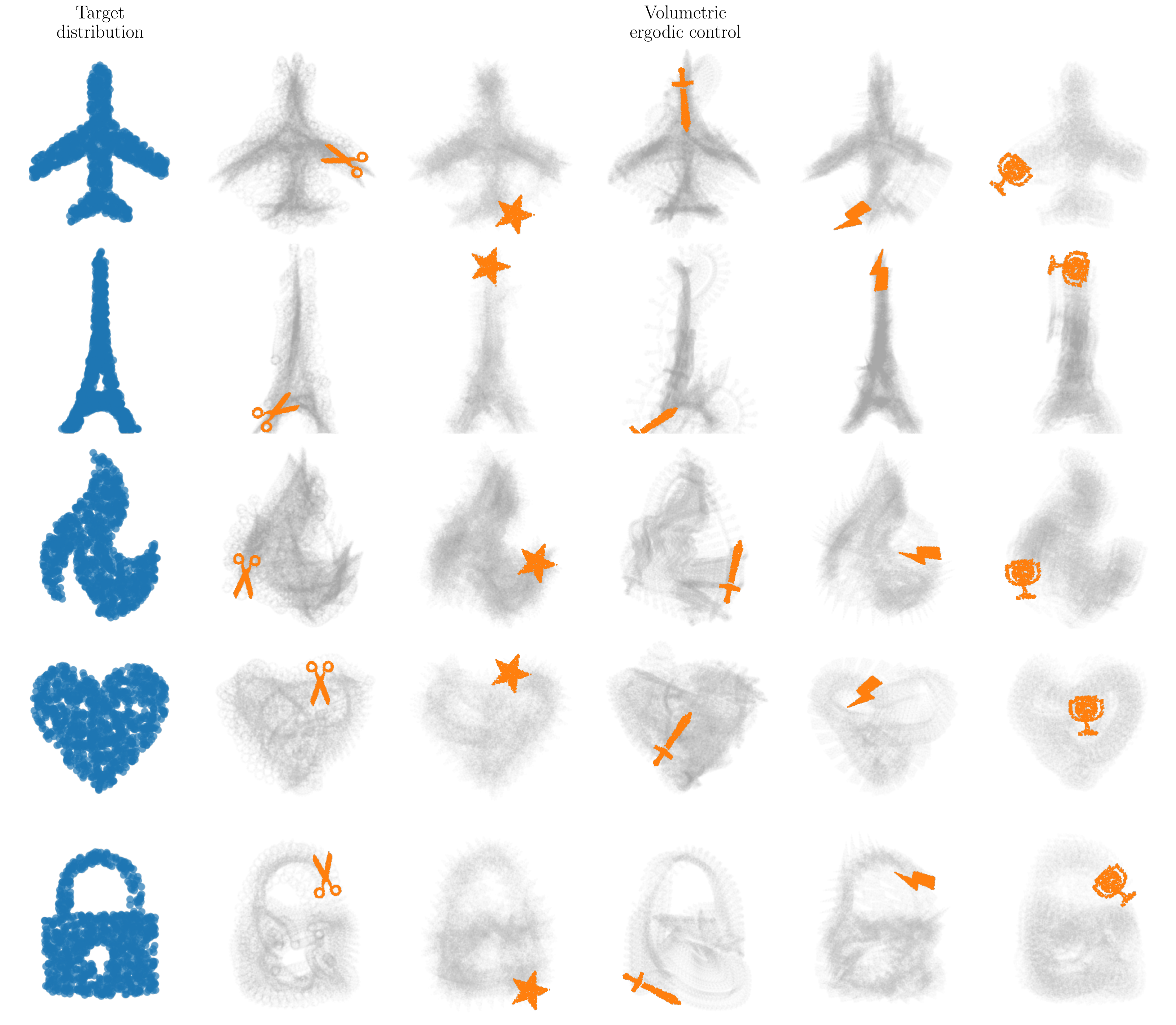}
    \vspace{-2em}
    \caption{\textbf{Erasing results using volumetric ergodic control (VEC).} The first column shows the target distributions (\emph{airplane, Eiffel Tower, fire, heart, lock}) and remaining columns show VEC applied with different end-effector geometries (\emph{scissors, star, sword, lightning, trophy}), which are modeled as volumetric states with the sample-based representation. VEC reasons over the volumetric states and produces efficient and thorough coverage trajectories.}
    \label{fig:erasing_final_timestep}
    \vspace{-1em}
\end{figure*}

\section{Introduction}

Ergodic control generates robot trajectories such that the time-averaged spatial statistics match a target distribution, ensuring that time spent in any region is proportional to the target density. This produces non-myopic and efficient coverage that prioritizes high-density regions while guaranteeing asymptotic coverage of the search space. Ergodic control has been shown to be effective across a wide range of tasks including autonomous information gathering~\cite{miller_ergodic_2016, Lee-RSS-24, dong_time_2025}, manipulation~\cite{shetty_ergodic_2022, bilaloglu_whole-body_2023, bilaloglu_tactile_2025}, and robot learning~\cite{prabhakar2022mechanical, pinosky_embodied_2024, berrueta_maximum_2024, SunM-RSS-25}. 

Many robotics tasks depend not only on position but on how the robot’s volumetric body and sensors interact with the world. In manipulation, tool and end-effector geometry are critical~\cite{cutkosky_grasp_1989, mahler_dex-net_2017, ha_fit2form_2021}. In autonomous information gathering, the sensor’s field of view, range, and occlusions govern what can be observed and are essential for generating information-rich measurements~\cite{galceran_survey_2013, bouman_adaptive_2022}. However, standard ergodic control methods model the robot as a non-volumetric point mass~\cite{mathew_metrics_2011,miller_ergodic_2016,sun_fast_2025}, ignoring the physical volume of the robot’s body or sensors that interact with the environment. While some extensions of ergodic control model the interaction beyond a point mass~\cite{ayvali_ergodic_2017,bilaloglu_whole-body_2023}, they remain limited to specific applications and do not generalize to a broader range of geometries and sensor models.

In this work, we extend the standard ergodic control formulation~\cite{mathew_metrics_2011} to incorporate volumetric state representations. We also introduce a sample-based volumetric representation that accommodates arbitrary robot bodies or sensor models, provided these models can be expressed as samples (see Fig.~\ref{fig:erasing_final_timestep} for examples). We call our method \emph{volumetric ergodic control (VEC)}. Our method preserves the ability of standard ergodic control to handle nonlinear dynamics and guarantee asymptotic coverage, while adding minimal computational overhead for real-time control. We benchmark VEC through a series of tasks against standard ergodic control, including erasing tasks with varying end-effector shapes and search tasks with a differential-drive robot and a quadcopter. Experimental results show that our method improves ergodic coverage efficiency by more than a factor of two while achieving a $100\%$ task completion rate across experiments. Lastly, we demonstrate the effectiveness of VEC on a Franka robot performing mechanical erasing tasks. In summary, the contributions of this paper are threefold:
\begin{enumerate}[leftmargin=*]
    \item Ergodic control formulation based on volumetric states;
    \item Demonstrating that VEC is compatible with and benefits from sample-based representations;
    \item Validation of VEC in simulation and hardware, demonstrating improved coverage efficiency and reliability.
\end{enumerate}

\section{Related Work}

\subsection{Ergodic control}

The standard ergodic control method is introduced in~\cite{mathew_metrics_2011}, which defines a Sobolev norm–based metric to evaluate the discrepancy between the spatial statistics of a trajectory and a target distribution, and proposes a receding-horizon control algorithm with an infinitesimal time horizon. This framework is extended in~\cite{miller_ergodic_2016} by applying the iterative linear quadratic regulator (iLQR) for control optimization. Since then, several new ergodic metric formulations and control optimization algorithms have been proposed to address limitations of the standard ergodic control, including improved optimization efficiency~\cite{mavrommati_real-time_2018}, multi-agent coordination~\cite{abraham_decentralized_2018}, scalability~\cite{shetty_ergodic_2022,sun_fast_2025}, and coverage on curved surfaces~\cite{dong_ergodic_2025, bilaloglu_tactile_2025}. 

A small number of extensions attempt to incorporate robot or sensor geometry, such as via kernels with Kullback-Leibler (KL) divergence~\cite{ayvali_ergodic_2017} or by decomposing manipulators into diffusion-driven agents~\cite{bilaloglu_whole-body_2023}. However, these methods often assume simple and limited robot and sensor geometry and are restricted to narrow applications. Our method fills this gap by directly incorporating a volumetric state representation into the formulation of the standard ergodic metric~\cite{mathew_metrics_2011}, producing more effective coverage trajectories while preserving the benefits of ergodic control.

\subsection{Volumetric representations in motion planning}

Beyond ergodic control, volumetric representations are essential in many other areas of motion planning. In manipulation, task performance depends on the geometry of tools and end-effectors, including their contact surfaces, orientations, and compliance~\cite{cutkosky_grasp_1989, mahler_dex-net_2017}. In autonomous information gathering, the sensor’s field of view, which varies with altitude and attitude and is often anisotropic, affects what can be observed and what information is collected~\cite{galceran_survey_2013, bouman_adaptive_2022}. In structural inspection and environmental monitoring, feasible coverage is constrained by visibility and viewing geometry~\cite{englot_sampling-based_2012, hasan_oceanic_2024}. In agriculture and surface disinfection, coverage quality depends on per-pass width and per-area intensity, linking robot speed and altitude directly to task performance~\cite{martin_effect_2019}. These examples highlight the need for volumetric reasoning in ergodic control.

\section{Preliminaries on Ergodic Control}

\subsection{Notation}
Without loss of generality, we assume the search space, denoted as $\mathcal{X}$, is an $N$-dimensional rectangular space $\mathcal{X}=[0,L_1]\times\cdots\times[0,L_d]
\subset\mathbb{R}^d$, where $L_i$ is the length of the $i$-th dimension. We denote the continuous-time robot trajectory as $s:[0,T]\mapsto\mathcal{S}$, and the control trajectory as $u:[0,T]\mapsto\mathcal{U}$, where $\mathcal{S}\subset\mathbb{R}^{n}$ is the state space of the robot, and $\mathcal{U}\subset\mathbb{R}^{m}$ is the space of feasible control signals. We denote the continuous-time dynamics of the robot as $\dot{s}(t) = f(s(t), u(t))$. 

Note that the search space $\mathcal{X}$ and the robot state space $\mathcal{S}$ are not necessarily the same. For example, for an aerial robot conducting information gathering over a planar surface, the search space is 2-dimensional, while the state space of the robot is at least 6-dimensional. In such cases, there exists a differentiable mapping that maps the state of the robot to a point in the search space, such as extracting the $x$ and $y$ position of the aerial robot during control optimization. For simplicity, we assume the state space coincides with the search space in the standard ergodic control problem.

\subsection{Ergodic control}

The ergodic metric quantifies the discrepancy between the time-averaged spatial statistics of a trajectory---formulated as an empirical distribution---and the target distribution, denoted as $q:\mathcal{X}\mapsto\mathbb{R}_0^+$. 

\begin{definition}[Dirac delta function]
    A Dirac delta function is a generalized function~\cite{lighthill_introduction_1958}---not characterized by point-wise mapping but by its inner product property---defined as follows:
    \begin{align}
        \int f(x) \delta(x-s) dx = f(s), \quad \forall s\in\mathcal{X}, \forall f\in\mathcal{C}(\mathcal{X}),
    \end{align} where $\mathcal{C}(\mathcal{X})$ denotes the space of continuous functions over the domain $\mathcal{X}$. 
\end{definition}

\begin{definition}[Trajectory empirical distribution]
    The trajectory empirical distribution is defined as:
    \begin{align}
        p_s(x) = \frac{1}{T} \int_0^T \delta(x - s(t)) dt, \label{eq:emp_distr}
    \end{align} where $\delta$ is the Dirac delta function.
\end{definition}

Direct evaluation of the discrepancy between the trajectory empirical distribution (\ref{eq:emp_distr}) and the target distribution using standard methods such as the KL divergence is challenging due to the presence of the Dirac delta function in (\ref{eq:emp_distr}). To address this challenge, the standard ergodic metric introduced in~\cite{mathew_metrics_2011} is based on a Sobolev norm-based discrepancy measure which leverages the inner product property of the Dirac delta function. The Sobolev norm is formulated using the normalized Fourier basis functions as a set of orthonormal bases in the function space, which are defined below.

\begin{definition}[Normalized Fourier basis functions]
    A normalized Fourier basis function, characterized by a vector index of natural numbers $\mathbf{k}=[k_1,\dots,k_d]\in\mathbf{K}\subset\mathbb{C}^{d}$, is defined as:
    \begin{align}
        f_{\mathbf{k}}(x) = \frac{1}{h_{\mathbf{k}}} \prod_{i=1}^{d} \cos\!\left( \frac{k_i\pi}{L_i}\, x_i \right), \label{eq:fourier}
    \end{align} where $h_{\mathbf{k}}$ ensures that $\int f_{\mathbf{k}}(x)^2 dx = 1$.
\end{definition}

Since the normalized Fourier basis functions form a set of orthonormal bases, we can decompose the probability density function of the target distribution $q(x)$ as follows:
\begin{gather}
    q(x) = \sum_{\mathbf{k}\in\mathbf{K}} \phi_{\mathbf{k}} f_{\mathbf{k}}(x) , \quad \phi_{\mathbf{k}} = \int q(x) f_{\mathbf{k}}(x) dx, 
\end{gather} where $\phi_{\mathbf{k}}$ is the Fourier coefficient of $q(x)$. While the exact orthonormal decomposition requires an infinite number of Fourier basis functions, in practice, we can approximate the decomposition with a finite number of Fourier coefficients.

More importantly, we can decompose the trajectory empirical distribution in the same way:
\begin{align}
    p_s(x) = \sum_{\mathbf{k}\in\mathbf{K}} c_{\mathbf{k}} f_{\mathbf{k}}(x) , \quad c_{\mathbf{k}} = \int p_s(x) f_{\mathbf{k}}(x) dx,
\end{align} where the inner product property of the Dirac delta function leads to closed-form evaluation of the Fourier coefficients:
\begin{align}
    c_{\mathbf{k}} & = \int \left[ \frac{1}{T} \int_0^T \delta(x - s(t)) dt \right] f_{\mathbf{k}}(x) dx \nonumber \\
    & = \frac{1}{T} \int_0^T \left[ \int f_{\mathbf{k}}(x) \delta(x - s(t)) dx \right] dt \nonumber \\
    & = \frac{1}{T} \int_0^T f_{\mathbf{k}}(s(t)) dt. \label{eq:fourier_coef}
\end{align}

Based on the Fourier coefficients, the standard ergodic metric is defined as follows.

\begin{definition}[Standard ergodic metric]
    The standard ergodic metric between a trajectory $s(t)$ and a target distribution $q(x)$ is defined as:
    \begin{align}
        \mathcal{E}(s(t), q(x)) = \sum_{\mathbf{k}\in\mathbf{K}} \lambda_{\mathbf{k}} \big(c_{\mathbf{k}} - \phi_{\mathbf{k}}\big)^2, \label{eq:erg_metric}
    \end{align} where $\{\lambda_{\mathbf{k}}\}$ is a convergent sequence:
    \begin{align}
        \lambda_{\mathbf{k}} = \left( 1 + \Vert \mathbf{k} \Vert^2 \right)^{-\frac{d+1}{2}},
    \end{align} which makes the formulation (\ref{eq:erg_metric}) equivalent to a Sobolev norm with a negative index of $\frac{1+d}{2}$.
\end{definition}

Ergodic control addresses the control optimization problem of minimizing the ergodic metric.

\begin{definition}[Standard ergodic control]
    \begin{gather}
        u(t)^* = \argmin_{u(t)} \mathcal{E}(s(t), q(x)), \label{eq:erg_ctrl} \\
        \text{s.t. } s(t) = s_0 + \int_0^{t} f(s(t), u(t)) dt , \quad \forall t\in[0, T].
    \end{gather}
\end{definition}

The optimization problem (\ref{eq:erg_ctrl}) can be solved through standard optimal control methods, such as nonlinear programming, iLQR, and augmented Lagrange multiplier. 

\section{Volumetric Ergodic Control}

\subsection{Volumetric state representation}

Standard ergodic control formulations model the robot as a non-volumetric point, neglecting the fact that in practice, the robot interacts with the environment through its body or a sensor, both of which occupy physical volume. 

\begin{definition}[Volumetric state representation]
    A volumetric state representation, defined as a state-dependent probability density function $g:\mathcal{X}\times\mathcal{S}\mapsto\mathbb{R}_0^+$:
    \begin{align}
        \int g(x, s) dx = 1, \quad \forall s\in\mathcal{S},
    \end{align} captures how the robot’s body or sensors, occupying physical volume, interact with the environment.
\end{definition}

\begin{definition}[Volumetric empirical distribution]
    Given a robot trajectory $s(t)$ with time horizon $[0,T]$ and the volumetric state representation $g(x,s)$, the trajectory empirical distribution is defined as:
    \begin{align}
        p_{s}^{\text{v}}(x) = \frac{1}{T} \int_0^T g(x, s(t)) dt, 
    \end{align} which is the spatial distribution associated with a robot trajectory under the influence of the volumetric state representation. 
\end{definition}

\subsection{Volumetric ergodic control}

In addition, the Fourier coefficients used in the ergodic metric must be redefined to apply to this volumetric empirical distribution. We derive the Fourier coefficients for the volumetric empirical distribution below:
\begin{align}
    c_{\mathbf{k}}^{\text{v}} & = \int_{\mathcal{X}} f_{\mathbf{k}}(x) p_{s}^{\text{v}}(x) dx \\
    & = \int_{\mathcal{X}} f_{\mathbf{k}}(x) \left[ \frac{1}{T} \int_0^T g(x, s(t)) dt \right] dx \\
    & = \frac{1}{T} \int_0^T \left[ \int_{\mathcal{X}} f_{\mathbf{k}}(x) g(x, s(t)) dx \right] dt \\
    & = \frac{1}{T} \int_0^T f_{\mathbf{k}}^{\text{v}}(s(t)) dt. \label{eq:vol_fourier_coef}
\end{align} We call $f_{\mathbf{k}}^{\text{v}}(x)$ the \emph{volumetric Fourier basis function}. 

\begin{definition}[Volumetric Fourier basis]
    Given a volumetric state representation $g(s,x)$, the corresponding volumetric Fourier basis function is defined as:
    \begin{align}
        f_{\mathbf{k}}^{\text{v}}(s(t)) = \int f_{\mathbf{k}}(x) g(x,s(t)) dx = \mathbb{E}_{x\sim g(\cdot, s(t))}[f_{\mathbf{k}}(x)], \label{eq:vol_fourier}
    \end{align} which is the expected Fourier basis function with respect to the spatial distribution of the volumetric state representation.
\end{definition}

Comparing the Fourier coefficients with volumetric representation (\ref{eq:vol_fourier_coef}) with the standard Fourier coefficients in (\ref{eq:fourier_coef}), we can see that the Fourier coefficients in both cases are the time averages of the corresponding Fourier basis function over the state trajectory of the robot. We can now define the volumetric ergodic metric and the corresponding ergodic control problem.

\begin{definition}[Volumetric ergodic metric]
    Given a volumetric state representation, the corresponding ergodic metric between a trajectory $s(t)$ and a target distribution $q(x)$ is defined as:
    \begin{align}
        \mathcal{E}^{\text{v}}(s(t), q(x)) = \sum_{\mathbf{k}\in\mathbf{K}} \lambda_{\mathbf{k}} \big(c_{\mathbf{k}}^{\text{v}} - \phi_{\mathbf{k}}\big)^2, \label{eq:vol_erg_metric}
    \end{align}
\end{definition}

\begin{definition}[Volumetric ergodic control]
    Given a volumetric state representation, the corresponding ergodic control problem is defined as:
    \begin{gather}
        u(t)^* = \argmin_{u(t)} \mathcal{E}^{\text{v}}(s(t), q(x)), \label{eq:vol_erg_ctrl} \\
        \text{s.t. } s(t) = s_0 + \int_0^{t} f(s(t), u(t)) dt, \quad \forall t\in[0, T].
    \end{gather}
\end{definition}

As shown in (\ref{eq:vol_erg_metric}), the volumetric ergodic metric, and therefore the corresponding control optimization problem, shares the same structure as the standard ergodic control problem (\ref{eq:erg_metric}). Thus, any control optimization algorithm for the standard ergodic control can be directly applied to solve the volumetric ergodic control problem.

\begin{figure*}[t!] 
    \centering    
    \includegraphics[width=\textwidth, keepaspectratio]{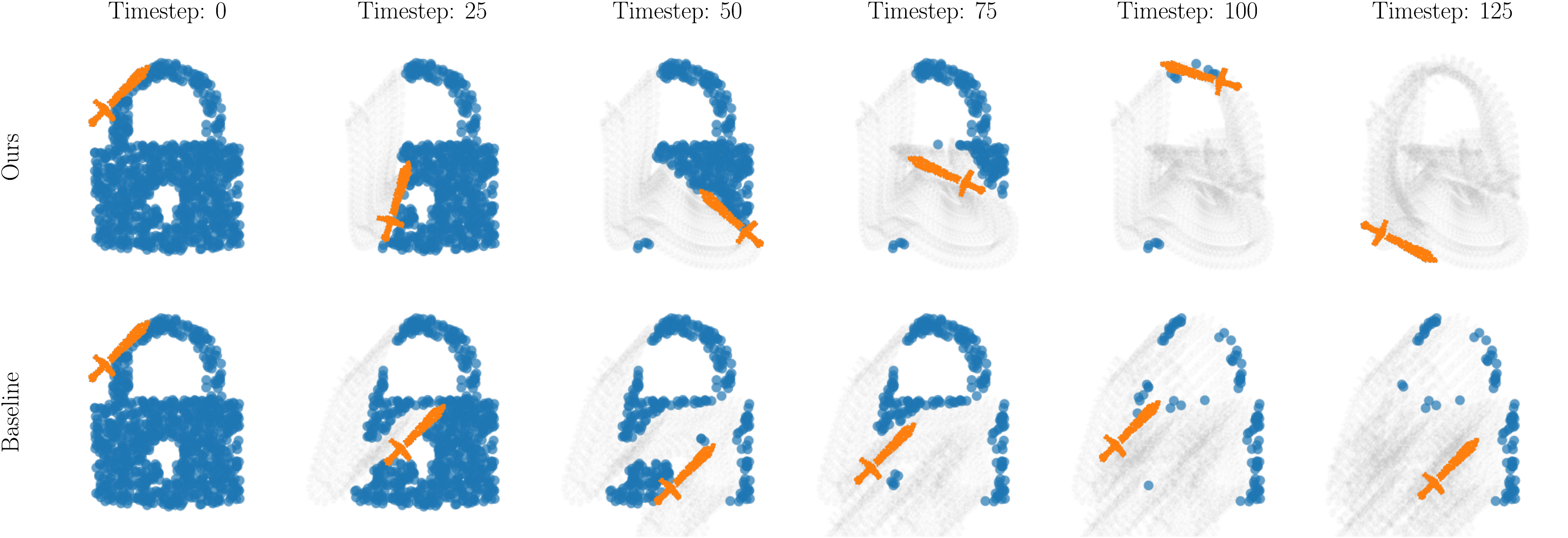}
    \vspace{-1em}
    \caption{\textbf{Qualitative results for erasing benchmark.} VEC effectively leverages the geometry of the end-effector through the volumetric state representation, generating both translational and rotational movements to completely erase the target in less time compared to the baseline ergodic control method.}
    \label{fig:erasing_047_sword_lock_panel}
\end{figure*}

\subsection{Sample-based volumetric representation}
\label{subsec:sample}

While the volumetric formulation above is defined in continuous space, in practice it is often more convenient to represent the robot’s volumetric state representation through a finite set of samples. Such sample-based volumetric representations naturally arise in applications such as ray-traced camera fields of view, end-effector contact points, or discretized body volumes. We approximate the state-dependent spatial density $g(\cdot,s)$ with an empirical measure induced by a finite set of samples that depend on the robot state $s$: 
\begin{align}
    g(x,s) = \frac{1}{N} \sum_{i=1}^{N} \delta(x-h_i(s)). \label{eq:sample_vol}
\end{align} Each sample is mapped from the robot state to the search space through a differentiable mapping $h_i: \mathcal{S}\mapsto\mathcal{X}$.

Substituting (\ref{eq:sample_vol}) into the definition of volumetric Fourier basis function (\ref{eq:vol_fourier}), the function can be expressed as:
\begin{align}
    f_{\mathbf{k}}^{\mathrm{v}}(s) & = \int f_{\mathbf{k}}(x) g(x,s(t)) dx \nonumber \\
    & = \int f_{\mathbf{k}}(x) \left[ \frac{1}{N} \sum_{i=1}^{N} \delta(x-h_i(s)) \right] dx \nonumber \\
    & = \frac{1}{N} \sum_{i=1}^{N} \int f_{\mathbf{k}}(x) \delta(x-h_i(s)) dx \nonumber \\
    & = \frac{1}{N} \sum_{i=1}^{N} f_{\mathbf{k}}(h_i(s)). \label{eq:fkv_sample_avg}
\end{align}

Sample-based representations of robot geometries and sensor models are widely used in robotics. For example, point cloud representations are common in manipulation~\cite{qi_pointnet_2017,wang_densefusion_2019,shridhar_perceiver-actor_2023}, while ray-casting and ray-tracing methods are standard in perception~\cite{mildenhall_nerf_2022,newcombe_kinectfusion_2011}. Importantly, in all such cases the resulting volumetric state representations are differentiable, which ensures compatibility with gradient-based control optimization. VEC shares the same computational complexity as standard ergodic control methods with respect to the number of Fourier coefficients, grid resolution, and planning horizon~\cite{sun_scale-invariant_2023, sun_fast_2025}. In addition, VEC scales linearly with the number of samples used to represent the volume. Lastly, VEC can leverage the same computational acceleration techniques developed for standard ergodic control, including parallelized computation~\cite{sun_scale-invariant_2023} and tensor-train decomposition~\cite{shetty_ergodic_2022}.

\subsection{Implementation details}

We apply the standard iterative linear quadratic regulator (iLQR) algorithm~\cite{miller_ergodic_2016} for control optimization, with the formulation also compatible with other control optimization methods, such as augmented Lagrange multiplier~\cite{dong_time_2025} and sequential action control~\cite{mavrommati_real-time_2018}. We implement the controller as a closed-loop receding-horizon controller based on state feedback, similar to the implementations in~\cite{mavrommati_real-time_2018,abraham_decentralized_2018}, where the robot re-plans the trajectory at each timestep based on all past states. Although the Fourier coefficients of the target distribution can be updated at each timestep in practice, we fix the target distribution in our benchmark evaluations to better assess the effectiveness of the volumetric state representation relative to the baseline. We implement our method in Python using the JAX library for numerical computation. 

\section{Experimental Results}
\subsection{Overview}

Our benchmarks address the following three questions:

\begin{enumerate}[label=Q\arabic*:]
    \item Do standard control optimization methods, such as receding-horizon iLQR, consistently reduce the proposed volumetric ergodic metric?
    \item Does volumetric state representation improve the performance of ergodic control with different end-effector geometries?
    \item Do volumetric state representations improve the performance of ergodic control in search tasks across different sensors and robot dynamics?
\end{enumerate}

\subsection{Experimental results for Q1}
\label{subsec:ergodicity}

\noindent\textbf{[Benchmark Q1 design]} We validate the effectiveness of VEC across three nonlinear dynamics and corresponding volumetric state representations:

\begin{enumerate}[leftmargin=*]
    \item Double-integrator: The volumetric state is a 2-D shape controlled from a randomly chosen pivot point.
    \item Differential-drive robot: The volumetric state is modeled as a forward-facing solid-state LiDAR.
    \item 12-DOF quadcopter: The volumetric state representation is obtained by ray-casting onto a plane.
\end{enumerate}

Across all three platforms, the task is to cover a randomized Gaussian mixture distribution. We test each platform for 25 trials with randomized initial states.

\begin{figure}[t!] 
    \centering    
    \includegraphics[width=0.45\textwidth, keepaspectratio]{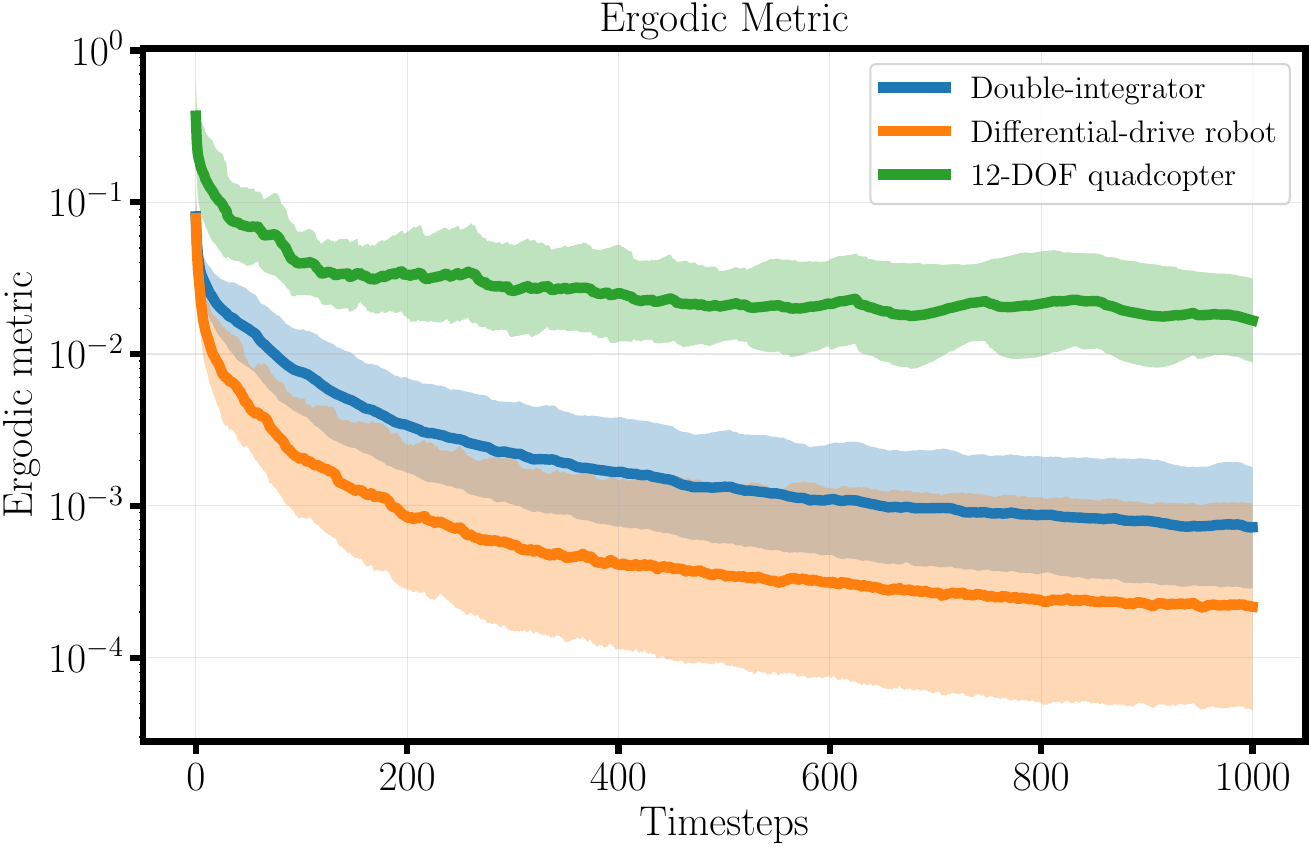}
    \vspace{-1em}
    \caption{\textbf{Validation of control optimization.} VEC is compatible with standard control optimization methods, such as iLQR. iLQR consistently minimizes the ergodic metric over time until convergence, across different robot dynamics, volumetric state representations, and randomized initial states. \emph{Note that the values of the converged ergodic metric depend on the specific robot dynamics and volumetric state representation, thus should not be directly compared across platforms.}}
    \label{fig:ergodicity_comparison}
    \vspace{-1em}
\end{figure}

\noindent\textbf{[Benchmark Q1 results]} As shown in Fig.~\ref{fig:ergodicity_comparison}, the iLQR consistently minimizes the volumetric ergodic metric across the three different platforms. These results demonstrate that the proposed volumetric ergodic control formulation is compatible with standard control optimization methods such as iLQR. Note that the converged ergodic metric value depends on the specific robot dynamics and the volumetric state representation, therefore should not be compared across robot platforms. Lastly, the average control frequencies are 8.3 Hz for the double-integrator, 9.5 Hz for the differential-drive robot, and 5.0 Hz for the 12-DOF quadcopter.

\begin{figure}[t!] 
    \centering    
    \includegraphics[width=0.45\textwidth, keepaspectratio]{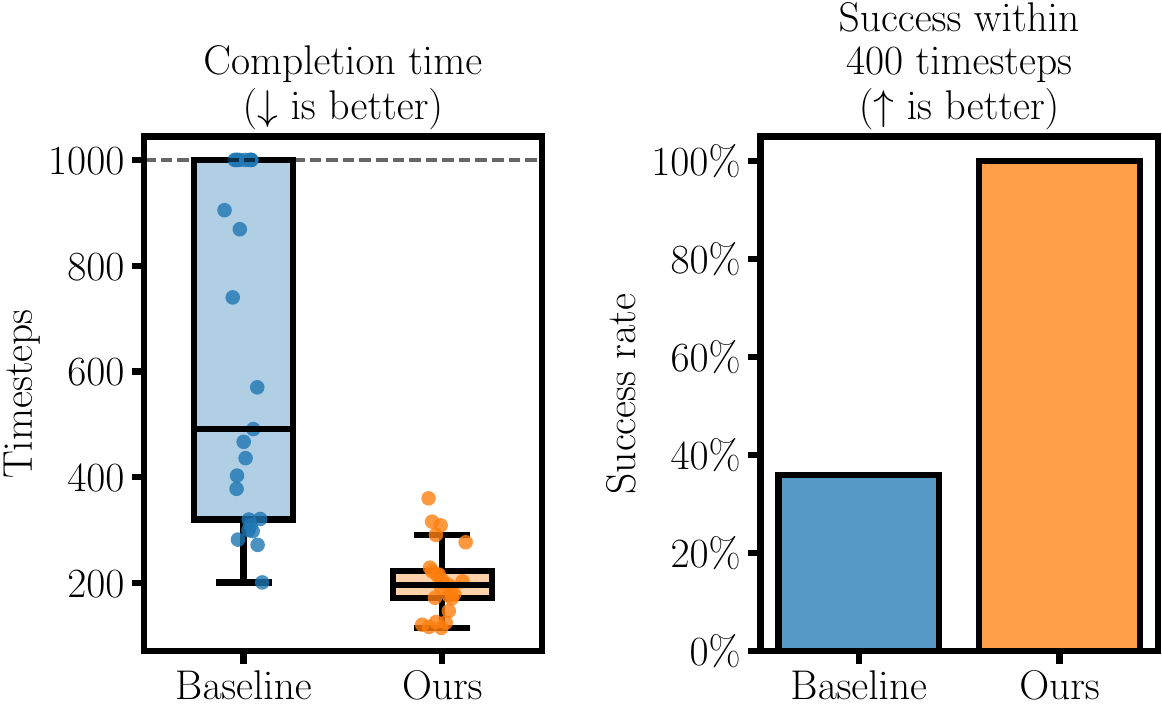} 
    \vspace{-1em}
    \caption{\textbf{Erasing benchmark.} (Left) VEC achieves $100\%$ task completion in less than half the time required by the baseline, which only completes $17$ out of the $25$ trials. (Right) VEC succeeds in all $25$ trials under 400 timesteps, while the baseline only completes 9 trials within the same amount of time.}
    \label{fig:erasing_both}

    \centering    
    \includegraphics[width=0.45\textwidth, keepaspectratio]{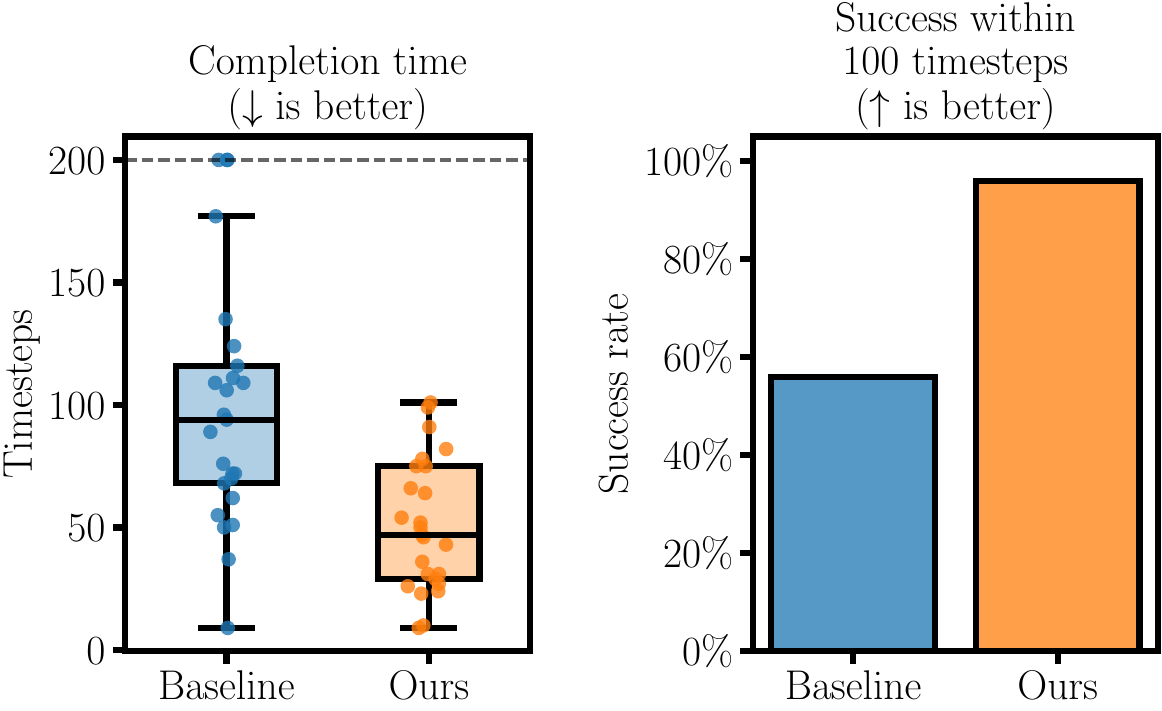}
    \vspace{-1em}
    \caption{\textbf{Ground search benchmark.}
    (Left) VEC achieves $100\%$ task completion with on average half as many timesteps as the baseline, which only completes $22$ out of the $25$ trials. (Right) VEC succeeds in $24/25$ trials under 100 timesteps, while the baseline only completes $14$ trials within the same amount of time.}
    \label{fig:diffdrive_second_order_both}
    
    \centering    
    \includegraphics[width=0.45\textwidth, keepaspectratio]{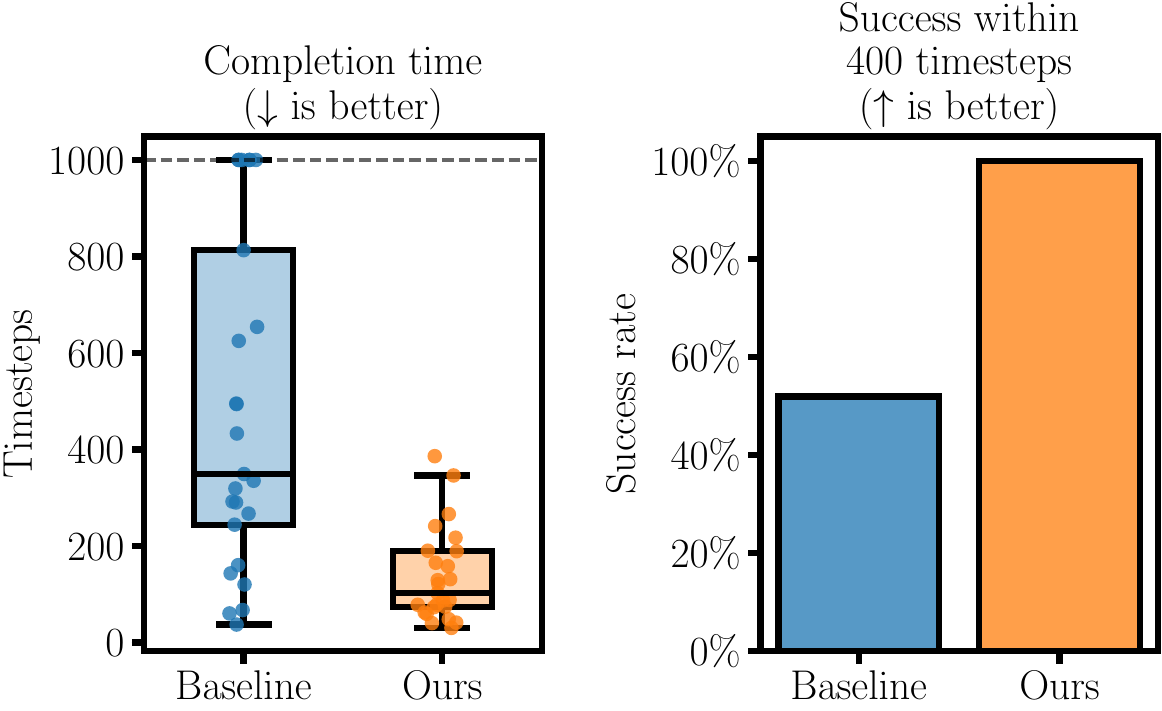} 
    \vspace{-1em}
    \caption{\textbf{Aerial search benchmark.} 
    (Left) VEC achieves $100\%$ task completion in less than half the time required by the baseline, which only completes $19$ out of the $25$ trials. (Right) VEC succeeds in all $25$ trials under 400 timesteps, while the baseline only completes 13 trials within the same amount of time.}
    \label{fig:quadcopter_both}
    \vspace{-1em}

\end{figure}

\begin{figure}[t!] 
    \centering    
    \includegraphics[width=0.45\textwidth, keepaspectratio]{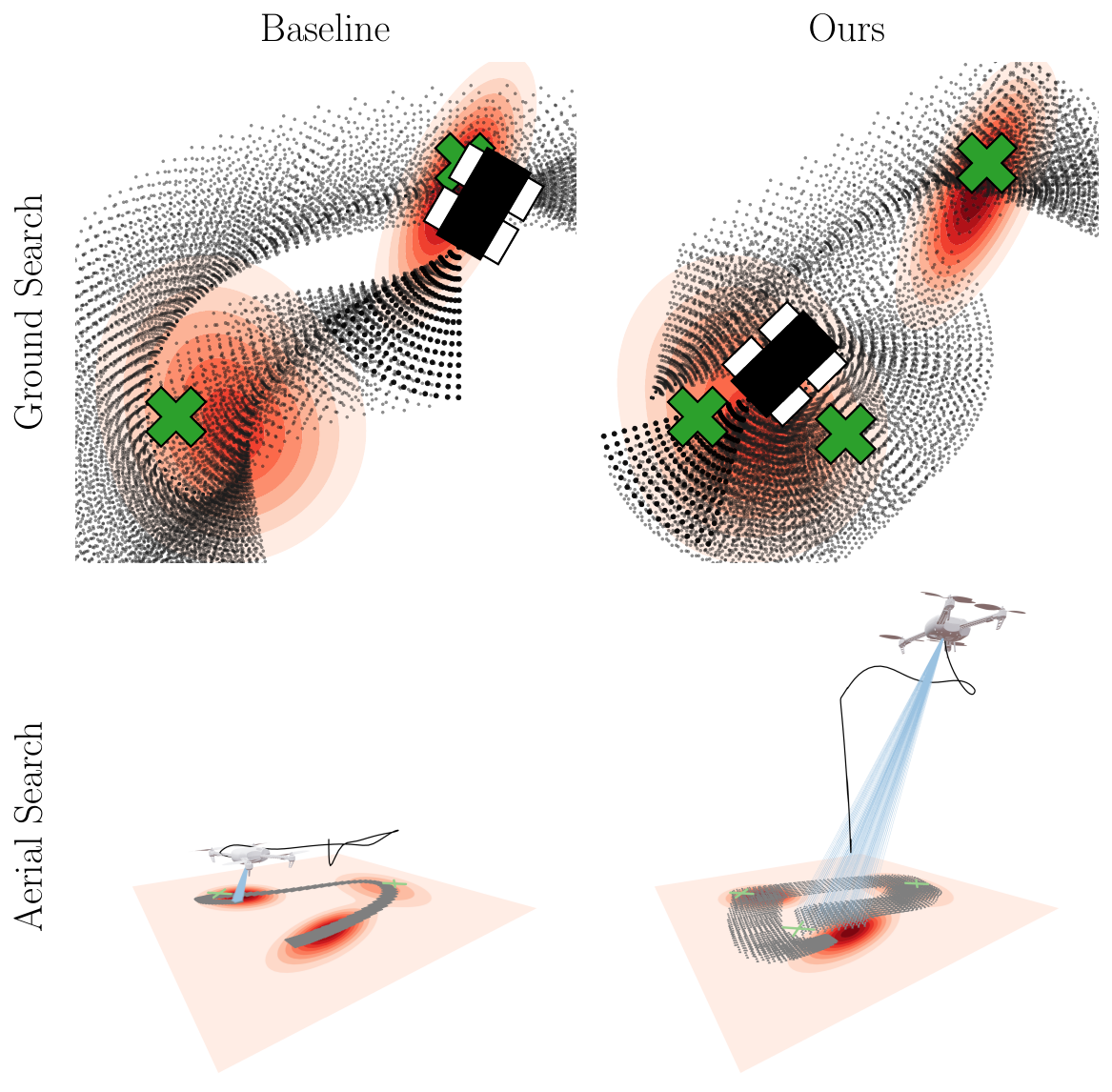}
    \vspace{-1em}
    \caption{\textbf{(Top) Qualitative results for ground search benchmark}. VEC leverages the volumetric LiDAR model to adapt the coverage trajectory, enabling faster and more reliable target discovery than the baseline ergodic control. \textbf{(Bottom) Qualitative results for aerial search benchmark}. VEC leverages the volumetric ray-casting camera model, enabling the quadcopter to ascend and widen its field of view, thereby completing the search more efficiently than baseline ergodic control, which remains at a constant height and performs sub-optimally.}
    \label{fig:ground_aerial_search}
    \vspace{-1em}
\end{figure}

\subsection{Experimental results for Q2}
\label{subsec:erasing}

\noindent\textbf{[Benchmark Q2 design]} We evaluate VEC on a 2-D erasing task, where the objective is to erase all points in the planar workspace. We use the five shapes---\emph{airplane, Eiffel Tower, fire, heart, lock}---as the target distributions and five tool geometries---\emph{scissors, star, sword, lightning, trophy}---as the volumetric state representation, all from the open-source benchmark in~\cite{SunM-RSS-25}. In each trial, we randomize the pivot point for controlling the tool, and the robot is modeled as a double integrator in its position and orientation. The baseline ergodic controller is implemented with the same dynamics and control optimization method, but is based on a point-based state representation at the pivot point. This ablation study focuses on the effect of volumetric state representation by keeping dynamics and control optimization identical across methods. 

\noindent\textbf{[Benchmark Q2 results]} Fig.~\ref{fig:erasing_047_sword_lock_panel} shows the qualitative results of a representative trial from the benchmark. For quantitative results, Fig.~\ref{fig:erasing_both} (Left) shows the timesteps required for task completion. VEC preserves the asymptotic coverage guarantee of ergodic control, achieving $100\%$ task completion ($25/25$ trials), whereas the baseline succeeds in only $17/25$ trials. Furthermore, the volumetric state representation significantly reduces the time required for task completion compared to the baseline ergodic control method. Fig.~\ref{fig:erasing_both} (Right) shows the task completion rate under a fixed budget of 400 timesteps. VEC succeeds in all $25/25$ trials within the time limit, while the baseline completes only $9/25$. This highlights the importance of volumetric state representation, which enables not only convergence but also efficient execution, ensuring reliable task completion in practice. Lastly, compared to the baseline, VEC adds only minimal computational overhead: the standard ergodic control method averages 87 ms per timestep, while VEC averages 132 ms despite using 1000 points for the volumetric state representation---both sufficient for real-time closed-loop control.

\begin{figure*}[t!] 
    \centering    
    \includegraphics[width=\textwidth, keepaspectratio]{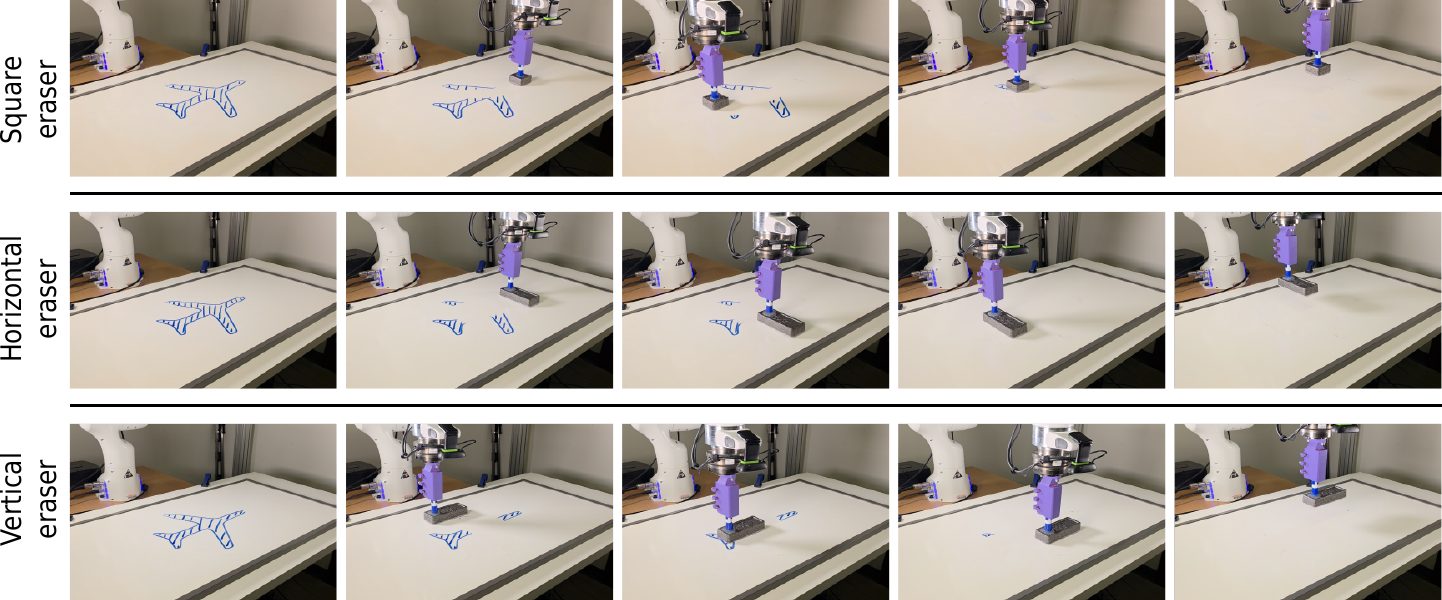}
    \vspace{-2em}
    \caption{\textbf{Franka erasing demonstration with multiple eraser shapes}. VEC accounts for the physical eraser size, producing smooth motions that efficiently erase the full target shape without redundancy. Snapshots show successful completion of the erasing task with multiple eraser geometries, including square (top), horizontal rectangle (middle), and vertical rectangle (bottom). More details are provided in the multimedia attachment.}
    \label{fig:franka_erasing}
    \vspace{-1em}
\end{figure*}

\subsection{Experimental results for Q3}
\label{subsec:search}

\noindent\textbf{[Benchmark Q3 design]} We evaluate VEC on two search tasks over a planar target distribution, where the objective is to find three unknown targets. Targets are sampled from a Gaussian mixture with fixed weights and randomized means and covariances. A target is considered found once any sensor measurement enters a fixed detection radius. The Gaussian mixture serves as a prior “signs of life” distribution indicating regions where targets are more likely to appear. The ground search task uses a second-order differential-drive robot with a forward-facing solid-state LiDAR. Uniform radial sampling yields a higher concentration of points near the robot, emphasizing local coverage. The aerial search task uses a 12-DOF quadcopter equipped with a ray-casting pinhole camera grid mounted at a $20^\circ$ forward tilt. Each pixel corresponds to a ray projected onto the ground plane, so the volumetric representation naturally expands, contracts, and skews with altitude and attitude. For both robots, the baseline ergodic controllers share the same dynamics and control optimization method, differing only in that they use point-based state representations. This design isolates the impact of volumetric state representation by holding dynamics and optimization identical across methods.

\noindent\textbf{[Benchmark Q3 results]} For ground search, qualitative results are shown in Fig.~\ref{fig:ground_aerial_search} (Top). Fig.~\ref{fig:diffdrive_second_order_both} (Left) shows the timesteps required for task completion. VEC preserves the asymptotic coverage guarantee of ergodic control, completing all $25/25$ trials compared to $22/25$ for the baseline, while also requiring on average only half as many timesteps. Fig.~\ref{fig:diffdrive_second_order_both} (Right) shows that under a 100-timestep budget, VEC succeeds in $24/25$ trials, whereas the baseline completes only $14/25$. Lastly, VEC adds minimal computational overhead: baseline averages 158 ms per timestep, while VEC averages 240 ms with 1000 points used as the volumetric state---both sufficient for real-time closed-loop control.

The performance improvement with VEC is even more drastic in the aerial search benchmark. Qualitative results are shown in  Fig.~\ref{fig:ground_aerial_search} (Bottom). Fig.~\ref{fig:quadcopter_both} (Left) shows the timesteps required for task completion. VEC completes all $25/25$ trials compared to only $19/25$ for the baseline, while also requiring fewer than half as many timesteps. Fig.~\ref{fig:quadcopter_both} (Right) shows that under a 400-timestep budget, VEC succeeds in all $25/25$ trials, whereas the baseline completes only $13/25$. Similarly, VEC adds minimal computational overhead: baseline averages 213 ms per timestep, while VEC averages 276 ms per timestep with 1000 points used as the volumetric state---both remaining within real-time closed-loop requirements.

\subsection{Erasing Demonstration with a Franka Robot}

We further demonstrate the effectiveness of VEC on a Franka robot for a series of erasing tasks with multiple end-effector geometries. The target distribution is taken from the numerical erasing benchmark and hand drawn on a whiteboard. VEC is then executed using the robot’s physical eraser geometry as the volumetric state. Fig.~\ref{fig:franka_erasing} shows snapshots from three representative experiments, with videos of the full experiments provided in the multimedia attachment. VEC achieves complete erasing in all experiments. Notably, by incorporating volumetric state representation, the controller successfully alters the spatial patterns of the generated motion to accommodate the end-effector geometry, demonstrating the effectiveness of VEC in real-world conditions.

\section{Conclusions and Discussions}

In this work, we introduce volumetric ergodic control (VEC), an extension of ergodic control based on a volumetric state representation. By introducing the volumetric Fourier basis functions, VEC replaces the standard ergodic control formulation, preserves the asymptotic coverage guarantee of ergodic control, and is solvable using standard control optimization methods such as iLQR. We further show that VEC is compatible with sample-based volumetric representations, which enables arbitrary robot and sensor geometries to be represented as samples of volumetric states. Comprehensive benchmark evaluations show that VEC consistently outperforms the standard ergodic control baselines across multiple robot types and their associated geometries or sensor models, achieving a $100\%$ task success rate across all benchmarks while introducing minimal computational overhead for real-time closed-loop control. Hardware demonstrations on a Franka robot further showcase the efficacy of VEC in real-world conditions. Looking forward, we will extend VEC to more complex manipulation tasks---such as insertion, grasping, and active tactile perception---where reasoning over volumetric representations is necessary.


\bibliographystyle{IEEEtran} 
\bibliography{references} 

\end{document}